\documentclass[runningheads]{llncs}
\usepackage[T1]{fontenc}
\usepackage{graphicx}
\usepackage{amsmath,amssymb}
\usepackage{bm}
\usepackage{xcolor}
\usepackage{colortbl}
\usepackage{booktabs}
\usepackage{wrapfig}
\usepackage{multirow}
\usepackage{url}

\begin{document}
\title{Patch Pruning Strategy Based on Robust Statistical Measures of Attention Weight Diversity in Vision Transformers}
\titlerunning{Patch Pruning Strategy Based on the Diversity of Attention}
\author{Yuki Igaue \and Hiroaki Aizawa}
\authorrunning{Y. Igaue et al.}
\institute{Graduate School of Advanced Science and Engineering\\
Hiroshima University\\
Higashi-Hiroshima, Japan\\
\email{\{m241838,hiroaki-aizawa\}@hiroshima-u.ac.jp}}
\maketitle
\begin{abstract}
Multi-head self-attention is a distinctive feature extraction mechanism of vision transformers that computes pairwise relationships among all input patches, contributing significantly to their high performance. However, it is known to incur a quadratic computational complexity with respect to the number of patches. One promising approach to address this issue is patch pruning, which improves computational efficiency by identifying and removing redundant patches. In this work, we propose a patch pruning strategy that evaluates the importance of each patch based on the variance of attention weights across multiple attention heads. This approach is inspired by the design of multi-head self-attention, which aims to capture diverse attention patterns across different subspaces of feature representations. The proposed method can be easily applied during both training and inference, and achieves improved throughput while maintaining classification accuracy in scenarios such as fine-tuning with pre-trained models. In addition, we also found that using robust statistical measures, such as the median absolute deviation in place of variance, to assess patch importance can similarly lead to strong performance. Furthermore, by introducing overlapping patch embeddings, our method achieves better performance with comparable throughput to conventional approaches that utilize all patches.

\end{abstract}
\section{Introduction}
Vision Transformers (ViTs)~\cite{vit} are deep learning models designed for computer vision tasks. Despite their architectural departure from conventional Convolutional Neural Networks (CNNs), ViTs have demonstrated remarkable performance across a wide range of vision tasks, including image classification. Today, ViTs are widely adopted as foundational models in various applications.

In ViTs, the input image is divided into uniformly sized patches that are processed in parallel. Multi-head self-attention~\cite{transformer} serves as a feature extraction mechanism that learns the relationships among these patches. However, this attention computation, which is based on dot products, incurs a computational complexity proportional to the square of the number of patches.

One approach to addressing this challenge is pruning redundant patches that do not contribute significantly to the final prediction. In particular, numerous methods have been proposed that prune patches based on values computed within the multi-head self-attention mechanism, such as removing patches according to the magnitude of attention weights~\cite{evit} or merging patches based on the similarity of key vectors~\cite{tome}. However, the perspective of utilizing the diversity of attention weights computed across different attention heads as an indicator for patch pruning has not been sufficiently explored.

In this paper, we propose a patch pruning method that removes redundant patches with low contribution to the final prediction, based on the diversity of attention. Specifically, in multi-head self-attention, patches exhibiting large variability in attention values across different heads, which indicates more diverse attention representations, are retained as important patches, while those with low variability are pruned. 

Our main contributions are summarized as follows:
\begin{itemize}
  \item We propose a patch pruning method that can be applied to pretrained ViT models without modifying their architecture, achieving improvements in computational cost and throughput during both training and inference.
  \item In image classification tasks, the method effectively reduces the number of patches with minimal performance degradation, making it a practical approach for model lightweighting.
  \item By allowing patches to overlap, the method attains superior classification accuracy at FLOPs comparable to or lower than those without pruning.
\end{itemize}

\section{Related Work}
\subsection{Attention Mechanism}
The attention mechanism~\cite{attention} refers to the operation of retrieving values corresponding to keys that match a given query. When the query, key, and value are all derived from the same input sequence, the mechanism is specifically referred to as self-attention.

Single-head self-attention~\cite{transformer} is a form of self-attention. Given an input sequence $\bm{X}=\{{\bm{x}_{\rm class}}, \bm{x}_1, \dots, \bm{x}_N \}$ consisting of $D$-dimensional vectors, i.e., $\bm{x}_i \in \mathbb{R}^D$, three linear projections $\bm{W}_Q \in \mathbb{R}^{D \times D'}$, $\bm{W}_K \in \mathbb{R}^{D \times D'}$ and $\bm{W}_V \in \mathbb{R}^{D \times D'}$ are applied to generate the query $\bm{Q}=\{{\bm{q}_{\rm class}}, \bm{q}_1, \dots, \bm{q}_N \}$, key $\bm{K}=\{{\bm{k}_{\rm class}}, \bm{k}_1, \dots, \bm{k}_N \}$, and value $\bm{V}=\{{\bm{v}_{\rm class}}, \bm{v}_1, \dots, \bm{v}_N \}$, respectively. The $i$-th query, key, and value vectors $\bm{q}_i, \bm{k}_i, \bm{v}_i \in \mathbb{R}^{D’}$ are computed as follows:
\begin{align}
  \bm{q}_i &= \bm{x}_i\bm{W}_Q,  \notag \\
  \bm{k}_i &= \bm{x}_i\bm{W}_K,  \\
  \bm{v}_i &= \bm{x}_i\bm{W}_V, \notag
\end{align}
where $N$ is the number of patches. Then, the attention weight is obtained by computing the dot product between the query and key, followed by normalization using the softmax function. The vector of the attention weights from $i$-th patch with respect to all tokens (hereafter referred to as the attention vector) $\bm{a}_{i}$ is defined as follows:
\begin{align}\label{attn}
\bm{a}_{i}={\rm softmax}\left(\frac{\bm{q}_i\bm{K}^T}{\sqrt{D'}} \right).
\end{align}
This computation requires quadratic complexity with respect to the number of input patches, which leads to a significant increase in computational cost when dealing with high-resolution images. Let the output sequence be denoted by $\bm{Y}=\{{\bm{y}_{\rm class}}, \bm{y}_1, \dots, \bm{y}_N \}$, where each $\bm{y}_i \in \mathbb{R}^D$. The $i$-th output of the attention layer $\bm{y}_i$ is computed using a linear projection $\bm{W}_O \in \mathbb{R}^{D' \times D}$ as follows:
\begin{align}\label{output}
\bm{y}_{i}=\left( \sum_{j=1}^N \bm{a}_{i} \bm{v}_j\right)\bm{W}_O .
\end{align}

However, single-head self-attention tends to aggregate information into a single attention representation, making it less effective in capturing complex patterns. To address this limitation, multi-head self-attention was introduced, in which the query, key, and value are respectively divided into multiple heads along the feature dimension and , and single-head self-attention is performed in parallel, allowing the model to capture diverse attention patterns across multiple feature subspaces.

Nevertheless, in practice, it has been observed that different heads often attend to similar features, limiting the diversity of learned representations. To promote diversity among heads, various techniques have been proposed, such as introducing regularization terms in the loss function to encourage inter-head dissimilarity~\cite{dr,attndiversity}, or incorporating learnable parameters into the attention computation, as seen in re-attention mechanisms~\cite{deepvit}.
\subsection{Patch Pruning}
Various approaches have been proposed in the field of computer vision to reduce the substantial computational cost associated with attention computation. Among them, one prominent strategy is pruning redundant patches that do not contribute to the final prediction.

DynamicViT~\cite{dynamicvit} is a method that dynamically prunes redundant patches based on the input, enabling a reduction in FLOPs and an improvement in throughput while maintaining competitive accuracy. However, since it introduces learnable parameters to estimate the importance score of each patch, gradients must be propagated through all patches during training. As a result, although the method employs a masking mechanism based on Gumbel-Softmax~\cite{gumbelsoftmax} to suppress the influence of redundant patches, these patches cannot be completely removed during training.

Some methods have been proposed that allow for the complete removal of patches during both training and inference without introducing additional parameters, thereby reducing computational cost. As shown in equation\eqref{output}, the output of the $i$-th patch in the attention layer can be represented as a weighted sum of the value vectors, where the weights are given by the $i$-th attention vector. In EViT~\cite{evit}, based on this property and the observation that the class token in vision transformers tends to pay more attention to class-relevant object regions than to those in non-object regions~\cite{emproperties}, it is assumed that the $j$-th entry in the attention vector of the class token reflects the importance of the $j$-th patch. Therefore, the attention vector from the class token to all patches in multi-head self-attention is used as importance scores. Referring to equation\eqref{attn}, the importance score for the $h$-th head $\bm{a}_{\rm class}^{(h)} \in \mathbb{R}^{N}$ is calculated using the $h$-th query $\bm{Q}^{(h)}=\{{\bm{q}_{\rm class}^{(h)}}, \bm{q}_1^{(h)}, \dots, \bm{q}_N^{(h)} \}$, where each $\bm{q}^{(h)}_i \in \mathbb{R}^{D'/H}$ and $h$-th key $\bm{K}^{(h)}=\{{\bm{k}_{\rm class}^{(h)}}, \bm{k}_1^{(h)}, \dots, \bm{k}_N^{(h)} \}$, where each $\bm{k}^{(h)}_i \in \mathbb{R}^{D'/H}$ as follows:
\begin{align}\label{attnhead}
\bm{a}_{\rm class}^{(h)}={\rm softmax}\left(\frac{\bm{q}_{\rm class}^{(h)}{\bm{K}^{(h)}}^T}{\sqrt{D'/H}} \right).
\end{align}
where $N$ is the number of patches, $H$ is the number of heads. The final importance score $\bar{\bm{a}} \in \mathbb{R}^{N}$ is obtained by averaging the attention values across all heads, as follows:
\begin{align}\label{attnscore}
\bar{\bm{a}}=\frac{1}{H}\sum_{h=1}^H \bm{a}_{\rm class}^{(h)}.
\end{align}
However, the final importance score is obtained by averaging the attention weights across all heads, which may compromise the diversity of attention patterns captured by different heads.

In addition, methods such as ToMe~\cite{tome} employ lightweight matching algorithms to effectively reduce the number of patches by merging similar ones based on the cosine similarity between their Keys, thereby achieving both high accuracy and efficiency.

\begin{figure}[t]
\centering
\includegraphics[scale=0.37]{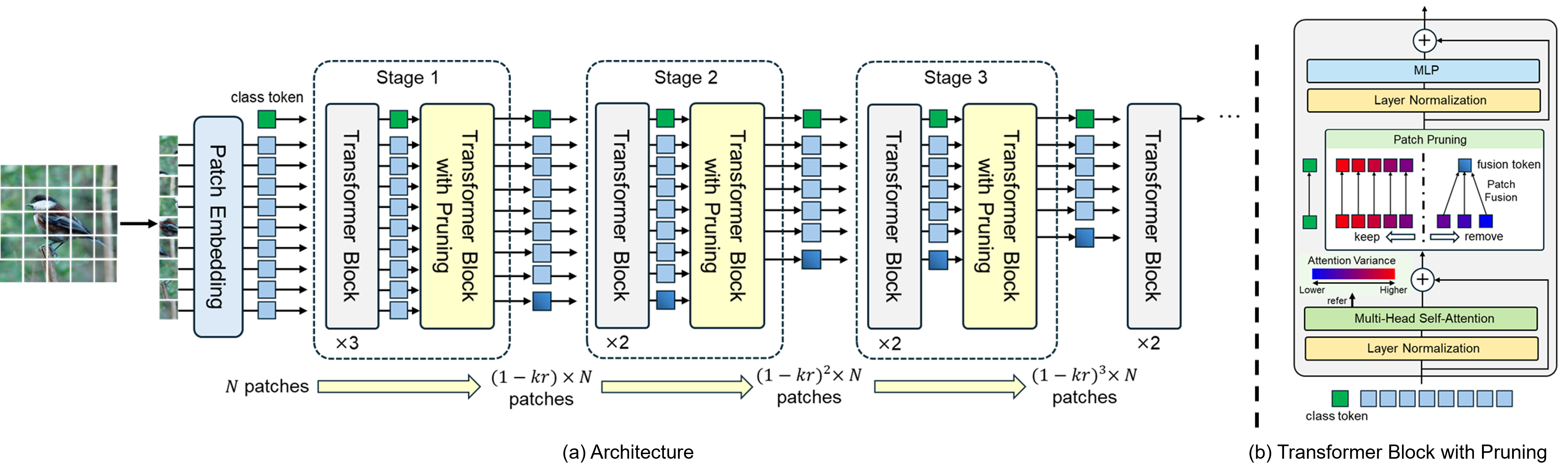}
\caption{(a) The architecture of a proposed method; (b) Transformer block with pruning.
}
\label{figure1}
\end{figure}

\section{Method}
\subsection{Overall Architecture}
An overview of the proposed architecture using a backbone network composed of 12 transformer blocks is illustrated in Fig.\ref{figure1}(a). Note that the backbone can employ various models, such as ViT~\cite{vit} and DeiT~\cite{deit}. In the proposed method, patch embeddings are first generated from the input image, and transformer blocks are then applied to these patches. Among the 12 blocks, the 4-th, 7-th, and 10-th blocks are replaced with transformer blocks with pruning, which are specifically designed for patch pruning. In this paper, we refer to a group consisting of multiple transformer blocks and one transformer block with pruning as a stage. At each stage, a keep rate $r \in (0,1)$ is defined to determine the proportion of patches to be retained. Since the discarded patches are not preserved in subsequent blocks, the computational cost is reduced progressively as the stages proceed.

As shown in Fig.\ref{figure1}(b), the layer normalization and multi-head self-attention are applied, and attention variance is computed for each patch based on the attention weights. The patches with low attention variance are identified as candidates for pruning. Note that these patches are fused into a single fusion token, which is appended to the end of the sequence instead of being entirely discarded. The details of attention variance and the patch fusion mechanism are described in the following section.

\subsection{Attention Weight Diversity based Patch Pruning Strategy}
\subsubsection{Pruning indicator}
Our hypothesis is that patches exhibiting diverse attention patterns across multiple heads are important in multi-head self-attention. Therefore, we compute the variance of attention weights across heads (referred to as attention variance) and regard patches with higher attention variance as more important. Specifically, following the assumption made in EViT~\cite{evit}, we consider that the attention weights from the class token to all patches indicate their importance. Based on this, we define attention variance $\bm{a}_{\rm var} \in \mathbb{R}^{N}$ as a pruning indicator by using the attention vector for the $h$-th head $\bm{a}_{\rm class}^{(h)}$ and the final importance score $\bar{\bm{a}}$ defined in equation\eqref{attnhead} and \eqref{attnscore} as follows:
\begin{align}
\bm{a}_{\rm var}=\frac{1}{H}\sum_{h=1}^H \left( \bm{a}_{\rm class}^{(h)}-\bar{\bm{a}} \right).
\end{align}

However, in the attention mechanism, although some tokens may receive excessively high attention weights, it has been reported that their influence is often canceled out through the matrix product between the attention weights and the value vectors~\cite{norm}. Based on this observation, we adopt the Median Absolute Deviation (MedAD) as an additional metric, which quantifies the dispersion of the distribution while minimizing the impact of outliers. we define attention MedAD $\bm{a}_{\rm MedAD} \in \mathbb{R}^{N}$ as another pruning indicator by using the attention vector for the $h$-th head $\bm{a}_{\rm class}^{(h)}$ in equation\eqref{attnhead} as follows:
\begin{align}
\bm{a}_{\rm MedAD}={\rm median} \left| \bm{a}_{\rm class}^{(h)}-{\rm median} \left( \bm{a}_{\rm class}^{(h)}\right)\right|.
\end{align}

\subsubsection{Patch fusion}
In the proposed method, instead of completely removing the pruned patches, they are aggregated into a single fusion token by taking a weighted sum based on the pruning indicator values and appended to the end of the sequence. This allows the model to preserve not only the information of the retained patches but also that of the pruned patches until the final layer. Furthermore, the softmax function is applied to scale the pruning indicator values $\bm{a}_{\rm var}$ or $\bm{a}_{\rm MedAD}$, depending on the configuration. We also introduce a temperature parameter $T \in (0,1)$ to emphasize the patches with higher indicator values. Let $\mathcal{P}$ denote the set of pruned patches and $a_j$ be the $j$-th element of the pruning indicator values, the fusion token $\bm{x}_{\rm fusion}$ is formulated as follows:
\begin{align}
\bm{x}_{\rm fusion} = \sum_{j \in \mathcal{P}} \frac{\exp(a_j / T)}{\sum_{k \in \mathcal{P}} \exp(a_k / T)} \cdot \bm{x}_j
\end{align}

\begin{table*}[t]
  \caption{Comparison of attention variance and attention MedAD}
  \label{table1}
  \centering
  \tiny
  \begin{tabular}{lc||ccc|ccc}
    \hline
    & & \multicolumn{3}{c}{w/o Patch Fusion} & \multicolumn{3}{c}{w/ Patch Fusion}\\
    \hline
    Indicator & $r$ & acc(\%) & GFLOPs & images/s & acc(\%) & GFLOPs & images/s \\
    \hline \hline
    None & 1.0 & 86.74 & 4.61 & 1312.4 & 86.74 & 4.61 & 1312.4\\
    \hline
    \multirow{7}{*}{\shortstack{Attention\\Variance\\$\bm{a}_{\rm var}$}} & 0.9 & 86.54 \textcolor{blue}{(-0.20)} & 4.00 \textcolor{blue}{(-13.3\%)} & 1473.0 \textcolor{blue}{(+12.2\%)} & 86.32 \textcolor{blue}{(-0.42)} & 4.03 \textcolor{blue}{(-12.5\%)} & 1457.5 \textcolor{blue}{(+11.1\%)} \\
     & 0.8 & 86.06 \textcolor{blue}{(-0.68)} & 3.45 \textcolor{blue}{(-25.1\%)} & 1727.9 \textcolor{blue}{(+31.7\%)} & 86.24 \textcolor{blue}{(-0.50)} & 3.48 \textcolor{blue}{(-24.4\%)} & 1684.5 \textcolor{blue}{(+28.4\%)} \\
     & 0.7 & 86.10 \textcolor{blue}{(-0.64)} & 3.00 \textcolor{blue}{(-43.3\%)} & 1967.2 \textcolor{blue}{(+49.9\%)} & 86.40 \textcolor{blue}{(-0.34)} & 3.04 \textcolor{blue}{(-34.1\%)} & 1941.5 \textcolor{blue}{(+47.9\%)} \\
     & 0.6 & 85.20 \textcolor{blue}{(-1.54)} & 2.61 \textcolor{blue}{(-43.3\%)} & 2284.3 \textcolor{blue}{(+74.1\%)} & 85.90 \textcolor{blue}{(-0.84)} & 2.64 \textcolor{blue}{(-42.7\%)} & 2241.2 \textcolor{blue}{(+70.8\%)} \\
     & 0.5 & 84.88 \textcolor{blue}{(-1.86)} & 2.29 \textcolor{blue}{(-50.4\%)} & 2569.8 \textcolor{blue}{(+95.8\%)} & 85.20 \textcolor{blue}{(-1.54)} & 2.31 \textcolor{blue}{(-49.8\%)} & 2537.9 \textcolor{blue}{(+93.4\%)} \\
     & 0.4 & 84.02 \textcolor{blue}{(-2.72)} & 2.03 \textcolor{blue}{(-56.0\%)} & 2886.4 \textcolor{blue}{(+119.9\%)} & 84.14 \textcolor{blue}{(-2.60)} & 2.05 \textcolor{blue}{(-55.6\%)} & 2839.0 \textcolor{blue}{(+116.3\%)} \\
     & 0.3 & 83.42 \textcolor{blue}{(-3.32)} & 1.80 \textcolor{blue}{(-60.8\%)} & 3202.6 \textcolor{blue}{(+144.0\%)} & 83.56 \textcolor{blue}{(-3.18)} & 1.82 \textcolor{blue}{(-60.5\%)} & 3145.1 \textcolor{blue}{(+139.6\%)}\\
     \hline
    \multirow{7}{*}{\shortstack{Attention\\MedAD\\$\bm{a}_{\rm MedAD}$}} & 0.9 & 86.46 \textcolor{blue}{(-0.28)} & 4.00 \textcolor{blue}{(-13.3\%)} & 1489.6 \textcolor{blue}{(+13.5\%)} & 86.62 \textcolor{blue}{(-0.12)} & 4.03 \textcolor{blue}{(-12.5\%)} & 1497.9 \textcolor{blue}{(+14.1\%)} \\
     & 0.8 & 85.98 \textcolor{blue}{(-3.45)} & 3.45 \textcolor{blue}{(-25.1\%)} & 1737.9 \textcolor{blue}{(+32.4\%)} & 86.14 \textcolor{blue}{(-0.60)} & 3.48 \textcolor{blue}{(-24.4\%)} & 1716.5 \textcolor{blue}{(+30.8\%)} \\
     & 0.7 & 85.80 \textcolor{blue}{(-0.94)} & 3.00 \textcolor{blue}{(-34.8\%)} & 1978.8 \textcolor{blue}{(+49.5\%)} & 85.94 \textcolor{blue}{(-0.80)} & 3.04 \textcolor{blue}{(-34.1\%)} & 1962.1 \textcolor{blue}{(+49.5\%)} \\
     & 0.6 & 85.40 \textcolor{blue}{(-1.34)} & 2.61 \textcolor{blue}{(-43.3\%)} & 2286.2 \textcolor{blue}{(+74.2\%)} & 85.82 \textcolor{blue}{(-0.92)} & 2.64 \textcolor{blue}{(-42.7\%)} & 2264.2 \textcolor{blue}{(+72.5\%)} \\
     & 0.5 & 85.10 \textcolor{blue}{(-1.64)} & 2.29 \textcolor{blue}{(-50.4\%)} & 2581.7 \textcolor{blue}{(+96.7\%)} & 85.08 \textcolor{blue}{(-1.66)} & 2.31 \textcolor{blue}{(-49.8\%)} & 2550.9 \textcolor{blue}{(+94.4\%)} \\
     & 0.4 & 84.70 \textcolor{blue}{(-2.04)} & 2.03 \textcolor{blue}{(-56.0\%)} & 2906.1 \textcolor{blue}{(+121.4\%)} & 84.72 \textcolor{blue}{(-2.02)} & 2.05 \textcolor{blue}{(-55.6\%)} & 2855.9 \textcolor{blue}{(+117.6\%)} \\
     & 0.3 & 83.84 \textcolor{blue}{(-2.90)} & 1.80 \textcolor{blue}{(-60.8\%)} & 3222.2 \textcolor{blue}{(+145.5\%)} & 84.04 \textcolor{blue}{(-2.70)} & 1.82 \textcolor{blue}{(-60.5\%)} & 3170.4 \textcolor{blue}{(+141.6\%)} \\
     \hline
  \end{tabular}
\end{table*}

\section{Experiments}
\subsection{Experimental Setup}
All models were initialized with DeiT-S pretrained on ImageNet-1k~\cite{imagenet}, and fine-tuned for 100 epochs. The training was conducted on ImageNet-100~\cite{imagenet100} with 100 classes and approximately 13.5k images, using a batch size of 256 on a single NVIDIA RTX A6000 GPU. The learning rate was scheduled using a cosine learning rate scheduler~\cite{cosine} with an initial value of 0.001, and the optimizer used was AdamW~($\beta_1 = 0.9$, $\beta_2 = 0.999$, weight decay$=0.05$)~\cite{adamw}. During both training and inference, input images were resized to a resolution of $224 \times 224$. For data augmentation during training, we applied RandomCrop, RandomHorizontalFlip, Mixup~\cite{mixup}, CutMix~\cite{cutmix}, and RandomErasing~\cite{erasing}. In addition, model throughput was measured during inference under the same conditions. The number of FLOPs was measured using fvcore.

\begin{table}[t]
  \caption{Comparison with existing methods}
  \label{table2}
  \centering
  \small
  \begin{tabular}{l||cccc}
    \hline
    Method & acc(\%) & GFLOPs & images/s \\
    \hline \hline
    Proposed(Variance) & 85.90 & 2.64 & 2241.2 \\
    Proposed(MedAD) & 85.82 & 2.64 & 2264.2 \\
    EViT & 85.68 & 2.64 & 2259.3 \\
    ToMe & 85.36 & 2.71 & 2009.9 \\
    \hline
  \end{tabular}
\end{table}

\begin{table*}[t]
  \caption{Performance comparison using overlapped patches}
  \label{table3}
  \centering
  \small
  \begin{tabular}{lc||cccccc}
    \hline
    Indicator & $r$ & acc(\%) & GFLOPs & images/s \\
    \hline \hline
    None & 1.0 & 86.74 & 4.61 & 1312.4 \\
    \hline
    \multirow{7}{*}{\shortstack{Attention\\Variance\\$\bm{a}_{\rm var}$}} & 0.7 &  87.38\textcolor{blue}{(+0.64)} & 5.17 \textcolor{blue}{(+12.1\%)} & 1124.0 \textcolor{blue}{(-14.4\%)} \\
     & 0.6 & 87.16 \textcolor{blue}{(+0.42)} & 4.51 \textcolor{blue}{(-2.2\%)} & 1287.6 \textcolor{blue}{(-1.9\%)} \\
     & 0.5 & 86.76 \textcolor{blue}{(+0.02)} & 3.95 \textcolor{blue}{(-14.3\%)} & 1446.9 \textcolor{blue}{(+10.2\%)} \\
     & 0.4 & 86.14 \textcolor{blue}{(-0.60)} & 3.50 \textcolor{blue}{(-24.1\%)} & 1607.4 \textcolor{blue}{(+22.5\%)} \\
     & 0.3 & 85.36 \textcolor{blue}{(-1.38)} & 3.13 \textcolor{blue}{(-32.2\%)} & 1791.3 \textcolor{blue}{(+36.5\%)} \\
     & 0.2 & 84.24 \textcolor{blue}{(-2.50)} & 2.82 \textcolor{blue}{(-38.9\%)} & 1968.6 \textcolor{blue}{(+50.0\%)} \\
     & 0.1 & 83.28 \textcolor{blue}{(-3.46)} & 2.58 \textcolor{blue}{(-44.1\%)} & 2121.6 \textcolor{blue}{(+61.7\%)} \\
     \hline
    \multirow{7}{*}{\shortstack{Attention\\MedAD\\$\bm{a}_{\rm MedAD}$}} & 0.7 & 87.08 \textcolor{blue}{(+0.34)} & 5.17 \textcolor{blue}{(+12.1\%)} & 1111.9 \textcolor{blue}{(-15.3\%)} \\
     & 0.6 & 87.08 \textcolor{blue}{(+0.34)} & 4.51 \textcolor{blue}{(-2.2\%)} & 1274.1 \textcolor{blue}{(-2.9\%)} \\
     & 0.5 & 86.62 \textcolor{blue}{(-0.12)} & 3.95 \textcolor{blue}{(-14.3\%)} & 1425.4 \textcolor{blue}{(+8.6\%)} \\
     & 0.4 & 85.86 \textcolor{blue}{(-0.88)} & 3.50 \textcolor{blue}{(-24.1\%)} & 1587.8 \textcolor{blue}{(+21.0\%)} \\
     & 0.3 & 85.94 \textcolor{blue}{(-0.80)} & 3.13 \textcolor{blue}{(-32.2\%)} & 1780.5 \textcolor{blue}{(+35.7\%)} \\
     & 0.2 & 84.64 \textcolor{blue}{(-2.10)} & 2.82 \textcolor{blue}{(-38.9\%)} & 1950.7 \textcolor{blue}{(+48.6\%)} \\
     & 0.1 & 83.86 \textcolor{blue}{(-2.88)} & 2.58 \textcolor{blue}{(-44.1\%)} & 2099.2 \textcolor{blue}{(+59.9\%)} \\
     \hline
  \end{tabular}
\end{table*}

\subsection{Image Classification}
\subsubsection{Comparison of attention variance and attention MedAD}
We compared attention variance and attention MedAD with and without patch fusion in terms of the best score of validation top-1 accuracy (acc), FLOPs, and throughput (images/s), by varying the keep rate $r$ from 0.3 to 0.9 in increments of 0.1. The results are summarized in Table \ref{table1}. For most keep rates under both indicators, the accuracy with patch fusion outperformed that without patch fusion. Furthermore, in the case of attention variance with patch fusion at $r=0.7$, the throughput improved by approximately 50\%, while the accuracy dropped by only about 0.3\%. Moreover, attention MedAD maintained stable accuracy even at lower keep rates, achieving less than 3\% accuracy degradation even when over 90\% of the patches were pruned.

\subsubsection{Comparison with existing methods}
We also compared our proposed patch pruning and fusion based on attention variance and attention MedAD with existing pruning methods such as EViT and ToMe. To ensure a fair comparison, we adjusted the patch keep rate or the number of pruned patches per block in ToMe so that the overall FLOPs were approximately equal across all methods. The results are shown in Table~\ref{table2}. Our methods outperformed the existing methods in terms of accuracy for both attention variance and attention MedAD.

\begin{figure}[t]
\centering
\includegraphics[scale=0.4]{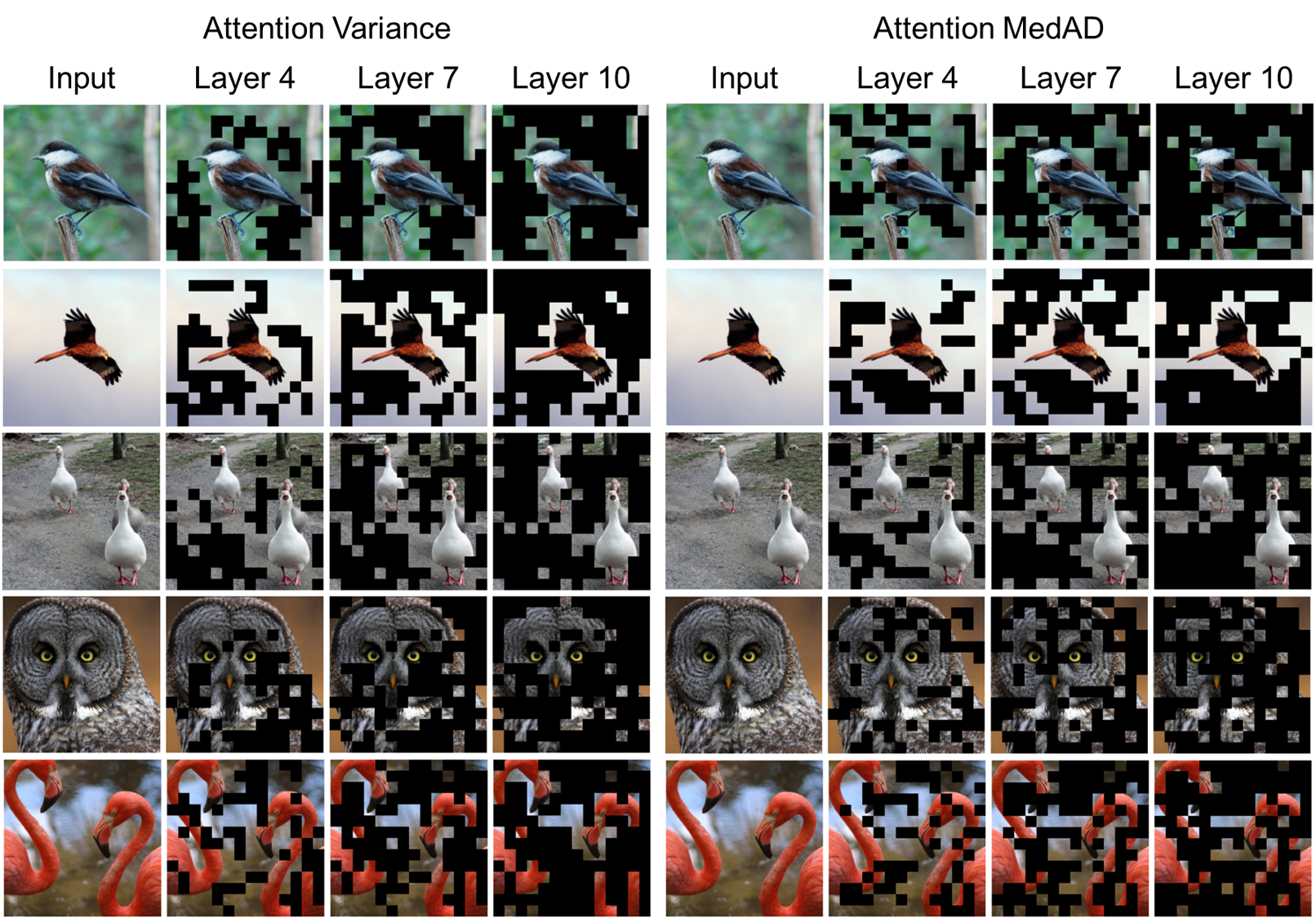}
\caption{\textbf{Comparison of Pruned Patches Across Transformer Blocks}}
\label{figure2}
\end{figure}

\begin{figure}[t]
\centering
\includegraphics[scale=0.5]{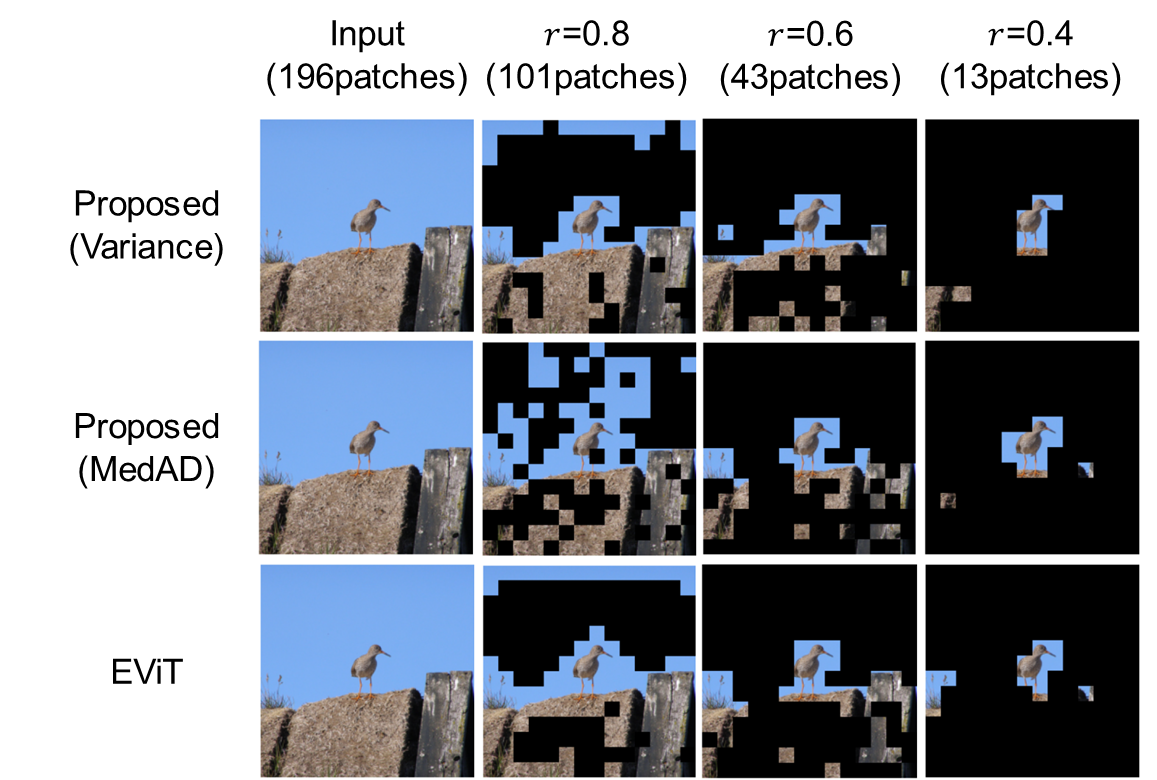}
\caption{\textbf{Comparison Between the Proposed Method and Existing Methods}}
\label{figure3}
\end{figure}

\subsubsection{With overlapped patches}
In conventional patch embedding, the input image is evenly divided without overlap by setting the stride equal to the patch size(=kernel size) in the Conv2d layer. In our approach, we intentionally introduce overlapping between patches by setting the stride smaller than the patch size(=kernel size). Specifically, we set the stride to patch size $\times \frac{3}{4}$, resulting in patches overlapping by $\frac{1}{4}$ of the patch size. This leads to redundant patch generation and increases the number of initial patches, which in turn raises the FLOPs. However, under low keep rates, pruning redundant patches allows for lower FLOPs compared to using non-overlapped patches without pruning.

we conducted comparisons in terms of the best score of validation top-1 accuracy (acc), FLOPs, and throughput (images/s) under the application of patch fusion for each pruning indicator by varying the keep rate from 0.1 to 0.7 in increments of 0.1. Table~\ref{table3} summarizes the results. In both attention variance and attention MedAD, the model achieved high accuracy at $r=0.6$, despite maintaining FLOPs and throughput comparable to the case without pruning. Moreover, it also demonstrated high accuracy at $r=0.5$, while achieving FLOPs and throughput equal to or lower than those at $r=0.9$ without patch overlapping.

\subsection{Visualization of Pruned Patches}
Figure~\ref{figure2} illustrates the pruned and retained patches at the 4th, 7th, and 10th Transformer blocks using the proposed attention variance and attention MedAD methods with a keep rate of $r=0.7$. In both methods, patches corresponding to background regions are mostly pruned. However, attention variance tends to retain more patches corresponding to the object regions until the final layer.

Figure~\ref{figure3} shows the retained patches at the final layer for the proposed methods and the baseline method (EViT) under keep rates of $r=0.4$, $0.6$, and $0.8$, respectively. For all keep rates, both our methods and EViT successfully preserve patches corresponding to the target objects.

\section{Conclusion}
we proposed a novel patch pruning method to address the ViTs challenge of quadratic computational complexity with respect to the number of patches in attention calculation. Our approach reduces patches based on a metric that captures diversity among attention heads in multi-head self-attention. The proposed method demonstrated superior top-1 accuracy compared to existing methods in fine-tuning for general object image classification tasks.

\section*{Acknowledgements}
This work was supported in part by JSPS KAKENHI Grant Number 25K21226.

\bibliographystyle{splncs04}
\bibliography{main}

\end{document}